# Enhanced Transformer Architecture for Natural Language Processing

**Woohyeon Moon, Taeyoung Kim, Bumgeun Park and Dongsoo Har**
Cho Chun Shik Graduate School of Mobility
Korea Advanced Institute of Science and Technology (KAIST)
Daejeon, South Korea moonstar,ngng9957,j4t123,dshar@kaist.ac.kr

## Abstract

Transformer is a state-of-the-art model in the field of natural language processing (NLP). Current NLP models primarily increase the number of transformers to improve processing performance. However, this technique requires a lot of training resources such as computing capacity. In this paper, a novel structure of Transformer is proposed. It is featured by full layer normalization, weighted residual connection, positional encoding exploiting reinforcement learning, and zero masked self-attention. The proposed Transformer model, which is called Enhanced Transformer, is validated by the bilingual evaluation understudy (BLEU) score obtained with the Multi30k translation dataset. As a result, the Enhanced Transformer achieves 202.96% higher BLEU score as compared to the original transformer with the translation dataset.

## 1 Introduction

The Transformer (Vaswani et al., 2017) has become a dominant model in natural language processing (NLP) (Zhao et al., 2023) and Computer Vision(Lai-Dang et al., 2023) revolutionizing numerous applications and establishing new performance benchmarks. This state-of-the-art Transformer model has paved the way for creating numerous large language models (LLM) (Lewis et al., 2019; Devlin et al., 2018) that are developed by stacking many Transformers.

By utilizing the Transformer architecture, these LLMs have reached unprecedented levels of language comprehension and generation (Lewis et al., 2019). They have proven useful in numerous NLP fields, such as machine translation (Lopez, 2008), text summarization (Tas and Kiyani, 2007), and question-answering (Kwiatkowski et al., 2019). Among these fields, the neural machine translation (NMT) (Bahdanau et al., 2014; Liu et al., 2020), which uses neural network (NN) (Kim et al., 2020) for machine translation, is being developed with globalization. The NMT model is now at the head of machine translation due to the capacity of the capturing long-range dependencies (Le et al., 2012) and contextual information (White et al., 2009).

It is widely believed that scaling up the model would directly enhance its performance. That is why most of the LLMs are based on scaling-up the model to improve the processing performance. However, this approach has inherent limitations, which includes rising training cost and computational inefficiency. Many of current works on the Transformer aim to improve the performance by investigating attention, decoder, etc (Leviathan et al., 2022; Zaheer et al., 2020; Kitaev et al., 2020; Ke et al., 2020a). However, these efforts only focus on the speed or improving limited part of Transformer architecture.

In this paper, we propose the Enhanced Transformer model as a novel solution to the aforementioned issues. Instead of focusing on training speed or only limited part of the Transformer, our proposed method strengthens four essential components of the original Transformer: layer normalization, residual connection, positional encoding, and the attention method.

The Enhanced Transformer overcomes the limitation of the conventional paradigm that has to increase the model size to achieve better performance by improving the performance of each component of the Transformer. Thus, the LLM developed from Enhanced Transformer is expected to be more efficient than the general LLM developed from the original Transformer, in terms of training cost. In this paper, the full layer normalization, weighted residual connection, positional encoding exploiting reinforcement learning (RL) (Vecchietti et al., 2020a,b), and zero masked self-attention techniques are applied to enhance the performance of the Transformer. The main contributions of this article are as follows.

1) Investigation of the structure and effect of

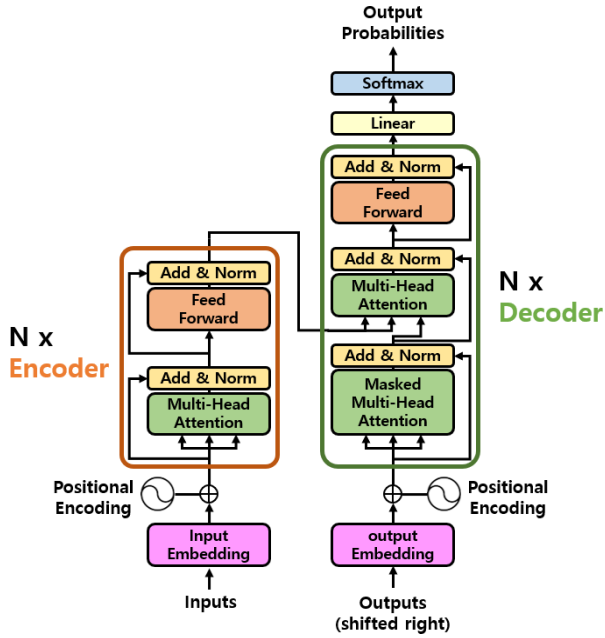

Figure 1: Structure of the Transformer

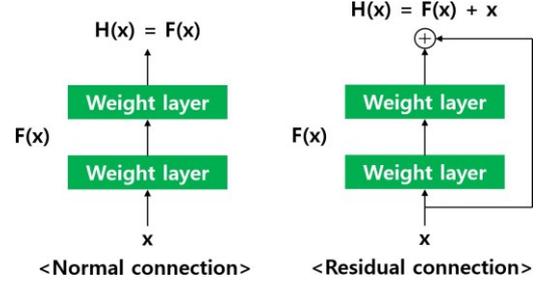

Figure 2: Structure of the residual connection. Left one is the normal connection method and the right one is the residual connection. Residual connection allows enhanced gradient flow, so the model with residual connection is resistant to the vanishing gradient problem.

normalization, residual connection, positional encoding, and attention method in the Transformer architecture.

2) Proposition of the new techniques full layer normalization, weighted residual connection, positional encoding exploiting RL, and zero masked self-attention for the improved performance of the Enhanced Transformer.

3) Development of effective LLM by improving the performance of the Transformer, which is the fundamental architecture of LLM for the various bilingual translation datasets.

The remainder of this paper is organized as follows. In Section 2, the structure of the Transformer is described. Section 3 presents the proposed Enhanced Transformer in details. Section 4 presents experimental results obtained with the original and Enhanced transformers. Concluding remarks are in Section 5. Some of translation results are listed in the Appendix.

## 2  Background

In this section, the most crucial components of the Transformer model, layer normalization, residual connection, positional encoding, and the attention method, are explained.

### 2.1  Layer normalization

In deep learning of NN, layer normalization (Ba et al., 2016) is a technique that is used to normalize the outputs of the activation functions within each layer. The layer normalization (Deatrick et al., 1999) is defined as:

$$y = \frac{x - E(x)}{\sqrt{V[x] + \epsilon}} \times \gamma + \beta \quad (1)$$

where the $x$ is the input value, the $E(x)$ is the mean value, the $V(x)$ is the variance value, the $\gamma$ and $\beta$ are the learnable parameters, and the $\epsilon$ is the minor constant preventing the denominator of equation (1) from taking a zero value, as added to the variance in the denominator.

### 2.2  Residual connection

Recently, the deep and large-scale NN model is widely employed. In the deep and large-scale NN model, the gradient of the model can either converge to zero or diverge to infinity as the number of layers increases; therefore, the vanishing gradient (Hochreiter, 1998) and exploding gradient (Philipp et al., 2017) issues can occur. To address these issues, the residual connection (Veit et al., 2016) method is developed. As shown in Figure 2, the residual connection adds the input value $x$ to the output value $F(x)$ to maintain the gradient value of the model, unlike the normal connection method, so the performance of the model is enhanced.

### 2.3  Positional Encoding

The greatest asset of the Transformer is the parallel arithmetic (von zur Gathen, 1986) which is more efficient than the series model, e.g., RNN family (Pascanu et al., 2013) in terms of computational speed and memory efficiency of GPU. The parallel arithmetic has a significant problem that the model does not know the location of each word token. To solve this problem, the concept of positional encoding (Vaswani et al., 2017) has been considered.

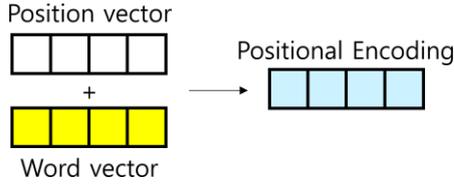

Figure 3: Structure of the positional encoding. The model adds the position vector to the word vector. Final output vector contains information about both the meaning and position of the word.

Positional encoding is the process of maintaining knowledge of the appearing order of words in a sequence. As shown in Figure 3, the positional vector, positional information of the word, is added to the word vector, which is the meaning of word. After the positional encoding process, the vector contains both the meaning and position information of word. The performance of the model is increased because the model knows the order of each word token, similar to series arithmetic. The positional encoding uses a sinusoidal function as follows:

$$PE_{(pos,2i)} = sin(pos/10000^{2i/d_{model}}) \quad (2)$$

$$PE_{(pos,2i+1)} = cos(pos/10000^{2i/d_{model}}) \quad (3)$$

where $pos$ represents the position of word, $i$ is the $i-th$ vector element of embedding, and $d_{model}$ is the size of embedding. The sinusoidal function offers several benefits in that the output is within the range of (-1~+1) and it allows sentences of any length due to its periodicity.

### 2.4 Attention

In RNN, the context vector is heavily influenced by the final RNN cell, so the information of the context vector can be lost. To resolve this issue, the attention mechanism (Vaswani et al., 2017; Zaheer et al., 2020; Kitaev et al., 2020) is proposed. The attention mechanism identifies the relationship between the words and determines which word receives massive correlation from the reference word. For instance, as shown in Figure 4, for the sentence "The animal did not cross the street because it was too tired", the attention mechanism calculates the meaning of 'it' and determines that 'it' is the 'animal'.

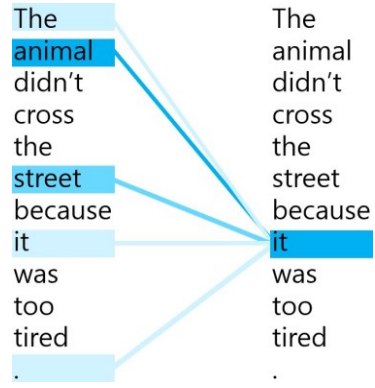

Figure 4: Example of the attention mechanism. 'it' is related with 'The', 'animal', 'street', 'it', '.' and 'animal' is the most related word among these.

The attention score is defined as a dot product of Query($Q$) and Key($K$). Because the mathematical meaning of the dot product is the degree of similarity, the dot product corresponds to the semantic similarity (Gardner et al., 2018). The length of the sentence also affects attention score and thus it is a significant issue. With the large length of $Q$ or $K$, the attention score can become enormous, regardless of the pure semantic similarity of words. It is not the original intent of the attention mechanism. To solve this problem, the dot product is divided by root $d_k$ for normalization. After passing it through the softmax activation function (Sharma et al., 2017), the attention value is finally obtained by multiplying a value $V$ as follows:

$$Attention(Q, K, V) = softmax\left(\frac{QK^T}{\sqrt{d^k}}\right)V \quad (4)$$

## 3 Proposed Method

In this section, we describe the problems of the original Transformer and introduce proposed model, Enhanced Transformer, that solves these problems. This section consists of 4 subsections, full layer normalization, weighted residual connection, positional encoding exploiting RL, and zero masked self-attention.

### 3.1 Full layer normalization

As suggested in Figure 1, the Transformer receives both input embedding and positional encoding as input data. In most cases, the input embedding and positional encoding have different data distributions, and that makes it difficult to train the model. To solve this challenging problem in training, a normalization layer is added in front of the original Transformer, and this technique is called here full layer normalization. The full layer normalization technique makes the distributions of input embedding and positional encoding similar. High performance of the model is obtained when the distribution of input datasets are similar to each other. Otherwise, the model performance tends to be degraded. From this viewpoint, the full layer normalization technique is used to add more normalization layers to the original Transformer for stable training of the model.

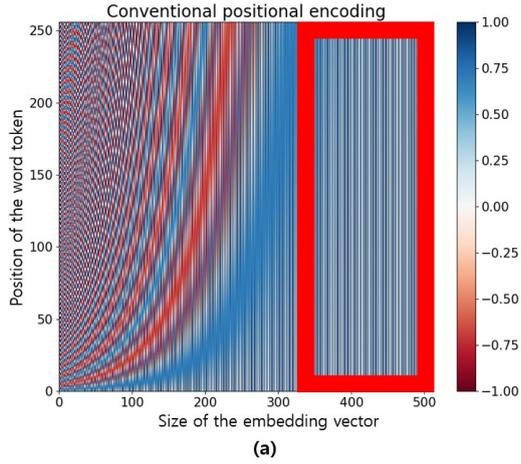

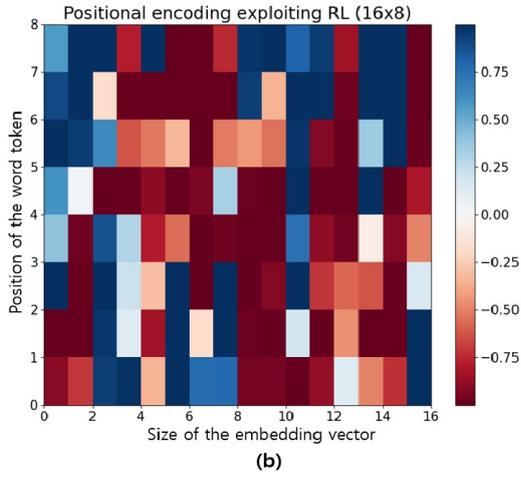

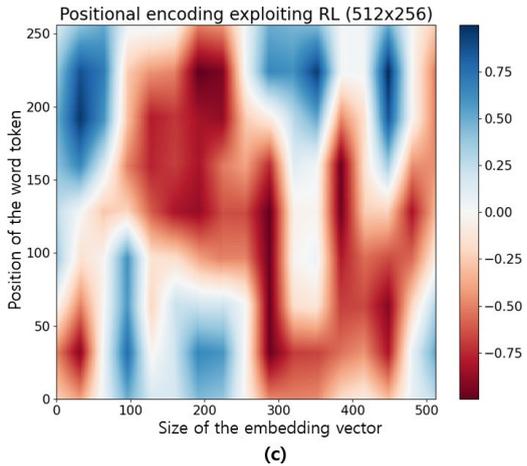

Figure 5: Visualized result of the positional encoding:(a) original positional encoding;(b) positional encoding exploiting RL without average pooling;(c) proposed positional encoding exploiting RL with average pooling. The proposed positional encoding is smoother than the other two including original one.

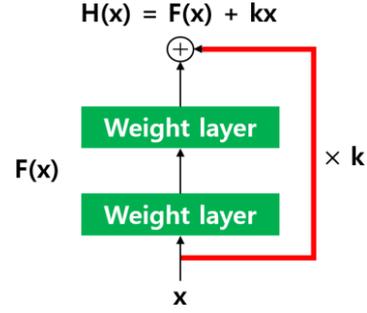

Figure 6: Weighted residual connection.

### 3.2 Weighted residual connection

In this subsection, the weighted residual connection that adds the weighted input value to the output value is suggested.

As shown in Figure 2, the original residual connection represents adding the first input vector $x$ to the output vector $F(x)$ as follows:

$$H(x) = F(x) + x \qquad (5)$$

In order to enhance gradient flow (Veit et al., 2016), the weighted residual connection is proposed. As shown in Figure 6, the weighted residual connection represents adding weighted input vector $kx$, instead of the original input vector $x$, to the output vector as follows:

$$H(x) = F(x) + kx \qquad (6)$$

where the $k$ is a hyperparameter similar to the learning rate. In subsection 4.3, we will find out the optimal value of $k$. The weighted residual connection enables loading a greater gradient, as the broader roadway can accommodate more traffic.

### 3.3 Positional encoding exploiting RL

Positional encoding is the process that adds the order of words to the word vector. In Figure 5,

the vertical axis of each subfigure represents the position of the word token and the horizontal axis represents the size of the embedding vector. As shown in the red box of Figure 5(a), each word vector at different positions is significantly identical in the original positional encoding which is according to equations (2) and (3). This observation is against the intent of employing positional encoding. To this end, positional encoding exploiting RL is proposed. RL technique is being employed for various applications (Moon et al., 2022).

In the proposed approach, the Soft Actor-Critic (SAC)(Haarnoja et al., 2018), one of the state-of-the-art models in RL, is used for positional encoding. This new positional encoding is designed to produce similar embeddings between word tokens in close positions and different embeddings between word tokens in distant positions. State of the SAC model is a 16 ×8 size vector as shown in Figure 5(b), and action of the SAC model is the assignment of scores between -1 to +1 to each element of state vector. And finally, the SAC agent gets reward from reward function as follows:

$$Reward = \frac{A \cdot B}{distance(A, B)} \quad (7)$$

where the reward is the similarity of words A and B divided by the distance between them. For every word pair of words, reward is calculated according to (7) and sum these rewards to assign to the agent. The agent of RL receives the highest reward when the similarity of words is close to '+1' with a small distance and the lowest reward when the similarity of words is close to '-1' with the small distance. The result of the SAC is shown in Figure 5(b). For training, the scaled-down state and action of SAC are used to save computing resources. Originally, the size of state× action is 512 × 256, which is too large for training of the SAC. For practical purpose, the scale-down state and action 16 × 8 is used, which is 1/1024(= $32^2$) times the original 512 × 256 state and action. Following this scaling down procedure, average pooling is applied to make a smooth positional encoding vector shown in Figure 5(c).

The new positional encoding using equation (7) for the reward function and shown in Figure 5(c) is significantly smoother than the one in Figure 5(b). The proposed positional encoding is 6.57 times better than the original positional encoding in the simulation environment score, which is taken by (7).

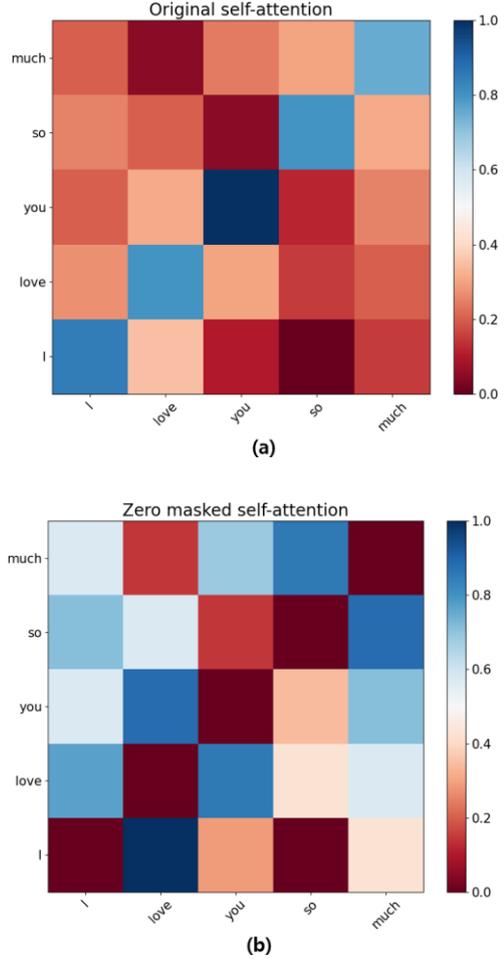

Figure 7: Heatmap of attention matrix :(a) original self-attention;(b) proposed zero self-attention. The proposed zero self-attention is seen to have more diverse scores than the original one.

### 3.4 Zero masked self-attention

As stated in Section 2, the attention mechanism of the Transformer is the most crucial component. Figure 7(a) shows the original self-attention scores. It is seen that the self-attention scores between the same words are distinctly higher . For instance, for the sentence "I love you so much," 'I' to 'I,' 'love' to 'love,' and 'you' to 'you' have significantly higher attention scores than the others because the most similar word to given one is itself. After normalization, the self-attention scores between a pair of different words become relatively small and thus hard to distinguish between words with small self-attention scores.

To solve this problem, the zero masked self-attention is suggested. The zero masked self-attention is a technique that sets the subdiagonal of an attention score matrix which is the highest score but least significant information, to zero. As shown in Figure 7(b), the zero masked self-attention generates more diverse distribution than the original one producing many similar scores.

## 4 Experimental Results

In this section, the experimental results of the Enhanced Transformer are presented and compared with those of other Transformer models already reported in the literature. In the first place, original model (Vaswani et al., 2017) is taken as the baseline model and each proposed technique of the Enhanced Transformer replaces corresponding one of the baseline model to measure the impact of it. Following this, performance of the entire Enhanced Transformer including all the proposed techniques is compared with other Transformers.

To evaluate the impact of individual proposed technique, the bilingual evaluation understudy (BLEU) score (Papineni et al., 2002) of German to English translation task is used. The 80 % of Multi30k-de-en dataset is used for training and remaining 20% is used for testing, In order to evaluate the performance of the entire Enhanced Transformer, the BLEU score of German to English, Korean to English, Chinese to English, and Japanese to English tasks, using 80 % of Multi30k datset for training and 20% for testing, are used. For training, learning rate = 1e-5, batch size = 128, maximal length = 256, dimension = 512, number of layers = 6, number of heads = 8, drop out rate = 0.2, and the epoch = 1000 episodes. Each training session lasts 24 hours.

### 4.1 Results of original Transformer

The performance of the original Transformer is evaluated in terms of the average BLEU score. The average BLEU score represents the average BLEU score of the latest 100 episodes. The blue curves in Figure 8 shows the results of the original model. The average BLEU score of the original model is 18.59.

### 4.2 Results of full layer normalization technique

The orange curve in Figure 8(a) represents the results of the model with full layer normalization technique. The average BLEU score of the model with the proposed technique is 42.31, which is 127.60% improved over the original model with original normalization.

### 4.3 Results of weighted residual connection technique

Figure 8(b) shows the results of the model with the weighted residual connection technique, where the orange, green, red, and purple curves indicate weights 2, 3, 4, and 5, respectively. As shown in the figure, 4 is the optimal weight for the full layer normalization technique. The average BLEU score of the model with the proposed technique is 29.70, which is 59.76% improved as compared to the original model.

### 4.4 Results of positional encoding exploiting RL technique

The orange curve in Figure 8(c) is the results of the positional encoding exploiting RL technique. The average BLEU score of the model with the positional encoding exploiting RL technique is 28.58, which is 53.74% improvement over the original model with original positional encoding.

### 4.5 Results of zero masked self-attention technique

The orange curve in Figure 8(d) represents the results of the original model with zero masked self-attention technique. The average BLEU score of the model with the proposed technique is 25.16, which represents 35.34% improvement.

### 4.6 Results of Enhanced Transformer

The Enhanced Transformer model is a model combined with full layer normalization, weighted residual connection, positional encoding exploiting RL, zero masked self-attention. As shown in Figure 9 and Table 1, the results of the Enhanced Transformer model is the orange curve. The average BLEU score of the Enhanced Transformer with Multi30k-de-en dataset is 56.32, and thus 202.96% improvement over the BLEU score of the original Transformer model. This number is larger than any improvement by any individually proposed technique.

The Enhanced Transformer is compared with other Transformer models, post layer normalization Transformer (Xiong et al., 2020), Realformer (He et al., 2020), and Transformer with untied positional encoding (Ke et al., 2020b). As shown in Table 1, the Enhanced Transformer demonstrates 20.82%, 46.97%, and 43.82% improvement over the post layer normalization Transformer, Realformer, and Transformer with untied positional en-coding, respectively.

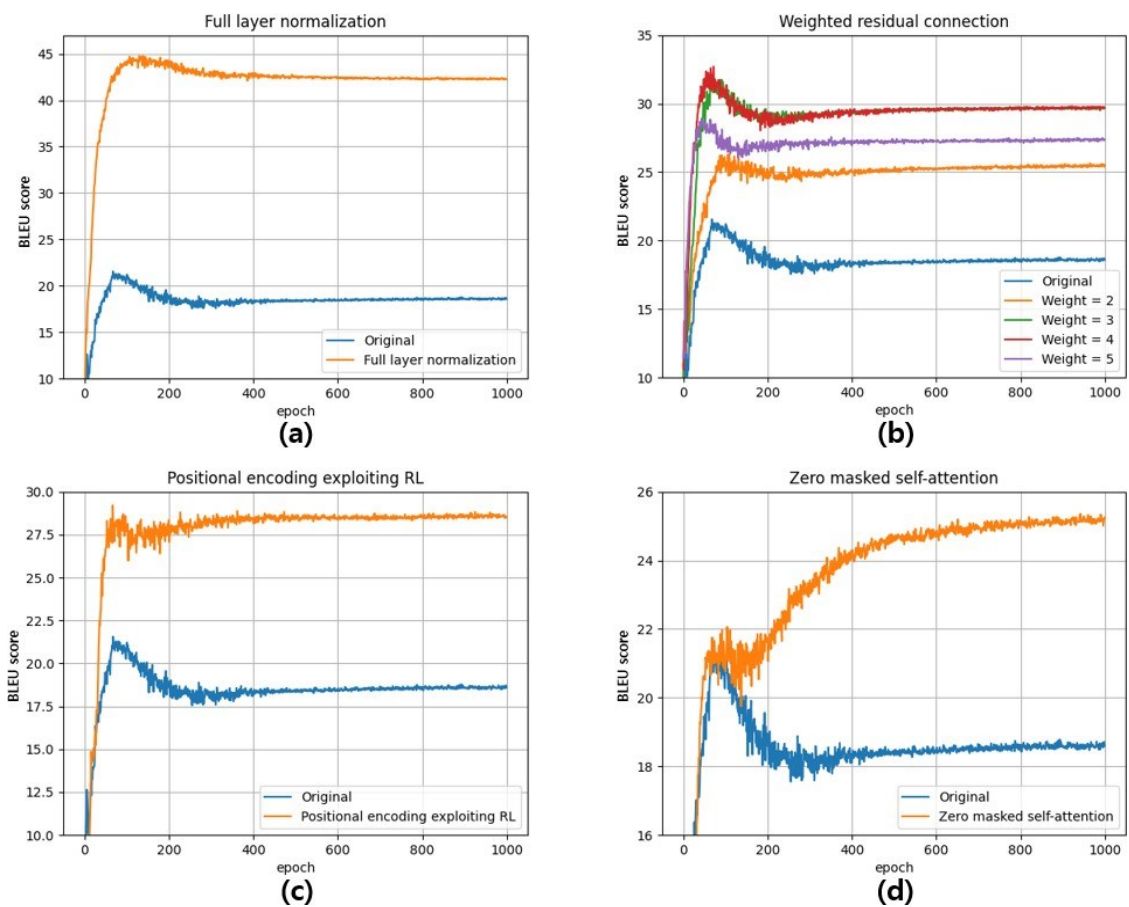

Figure 8: Performance of proposed techniques combined with original Transformer, using Multi30k-de-en dataset: (a) full layer normalization vs. original normalization ;(b) weighted residual connection vs. original residual connection;(c) positional encoding exploiting RL vs. original positional encoding;(d) zero masked self-attention vs. original self-attention. Blue curve represents performance of the Transformer with original technique and orange curve represents performance of the Transformer with proposed technique. Note different scales of vertical axes.

| Technique | Average BLEU score | Improvement(%) |
|---|---|---|
| Original | $18.59 \pm 5.43e(-2)$ | 0.00 |
| Full layer normalization | $42.31 \pm 4.86e(-2)$ | 127.60 |
| Weighted residual connection | $29.70 \pm 5.51e(-2)$ | 59.76 |
| Positional encoding exploiting RL | $28.58 \pm 7.42e(-2)$ | 53.74 |
| Zero masked self-attention | $25.16 \pm 5.98e(-2)$ | 35.34 |
| **Transformer model** | **Average BLEU score** | |
| Post layer normalization Transformer | $46.61 \pm 4.90e(-2)$ | |
| RealFormer | $38.32 \pm 5.27e(-2)$ | |
| Transformer with untied positional encoding | $39.16 \pm 4.38e(-2)$ | |
| Enhanced Transformer | $56.32 \pm 4.52e(-2)$ | |

Table 1: BLEU score of Transformer models. BLEU score of each technique represents that of original Transformer combined with it that replaces corresponding original technique. Multi30k-de-en dataset is used for comparison.

| Translation direction | Baseline | Proposed technique | Improvement(%) |
|---|---|---|---|
| German to English | $18.59 \pm 5.43e(-2)$ | $56.32 \pm 4.52e(-2)$ | 202.96 |
| Korean to English | $38.01 \pm 1.28e(-1)$ | $51.02 \pm 1.49e(-2)$ | 34.23 |
| Chinese to English | $65.39 \pm 5.00e(-2)$ | $87.82 \pm 1.41e(-14)$ | 34.30 |
| Japanese to English | $37.16 \pm 2.44e(-1)$ | $44.90 \pm 0.00$ | 20.83 |

Table 2: BLEU score of the proposed model with various translation directions. Multi30k dataset is used for comparison

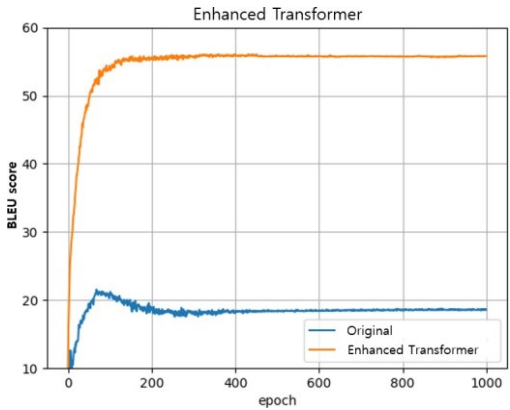

Figure 9: Performance of proposed Enhanced Transformer compared with that of original Transformer.

Until now, the results have been obtained with only Multi30k-de-en dataset. To prove that the Enhanced Transformer outperforms the original model generally, it is tested with other datasets in Multi30k dataset. They are datasets for Korean to English, Chinese to English, and Japanese to English translation. Table 2 shows the results of the Enhanced Transformer. As shown in Table 2, the Enhanced Transformer outperforms the original Transformer in general. Its improvement over the original Transformer is 34.23%, 34.30%, and 20.83% for Korean to English, Chinese to English, and Japanese to English translations, respectively. Also, the standard deviation of the BLEU score with the Enhanced Transformer is much smaller than that of the original one. This means that the proposed model achieves not only better performance but also better training stability. Examples of the translation results can be found in the appendix.

## 5 Conclusion

This paper introduces the Enhanced Transformer, a novel Transformer which enhances layer normalization, residual connection, positional encoding, and self-attention of the original Transformer. The proposed Enhanced Transformer is featured by four techniques: full layer normalization, weighted residual connection, positional encoding exploiting RL, and zero masked self-attention. The full layer normalization technique, weighted residual connection, positional encoding exploiting RL, and zero masked self-attention achieve improvement of 127.60%, 59.76%, 53.74%, and 35.34%, respectively, with Multi30k-de-en dataset. Entire Enhanced Transformer which is the combination of all these techniques achieves 73.08%(=(202.96+34.23+34.30+20.83)/4) average improvement over the original Transformer model with Multi30k dataset.

## Acknowledgements

This work was supported by the Institute for Information Communications Technology Promotion (IITP) grant funded by the Korean government (MSIT) (No.2020-0-00440, Development of Artificial Intelligence Technology that continuously improves itself as the situation changes in the real world).

# A Results of Translation

The translation result of our proposed model, Enhanced Transformer, is described. The baseline is the result of the original model, the proposed model is the result of the Enhanced Transformer, and the reference is the correct answer of translation.

| Translation direction | Baseline | Proposed technique | Reference |
|---|---|---|---|
| German to English | a young of and street of people woman is a is are a . a of | the picture has the mirror of a man in woman in on some woman . some sort of | the image shows the reflection of a man and woman located at a terminal of some sort |
| | a woman in a blue shirt and a shirt shirt and in a man white and | a woman with a red dress and red hair hair and in a bright colored colorful house | a woman in a red dress with red - hair standing before a bright , multicolored home |
| Korean to English | Maybe that is because there are few works of his his that have been stage into foreign what | Maybe that is because there are few works of his hers that have been translated into foreign languages | Maybe that is because there are few works of his hers that have been translated into foreign languages |
| | At an worried send I want to car a rateet so that I can listen to the rate | At an art gallery I want to borrow a headset so that I can listen to the explanations | At an art gallery I want to borrow a headset so that I can listen to the explanations |
| Chinese to English | a resolution authority may make one or more noise in petition in respect of a within within financial | a resolution authority may make one or more bail in instruments in respect of a within scope financial | a resolution authority may make one or more bail in instruments in respect of a within scope financial |
| | reasonable money in such manner on such terms and against such security as the board of medical thinks ex | borrow money in such manner on such terms and against such security as the board of trustees thinks ex | borrow money in such manner on such terms and against such security as the board of trustees thinks ex |
| Japanese to English | you don t have to get up so earlyughughughughion-ionionionionime | you don t have to get up so early | you don t have to get up so early |
| | he can speak both english and frenchughughughughion-ionionionionime | he can speak both english and french | he can speak both english and french |